\pdfoutput=1

\documentclass[11pt]{article}

\PassOptionsToPackage{table}{xcolor}
\usepackage[preprint]{acl}

\usepackage{times}
\usepackage{latexsym}

\usepackage[T1]{fontenc}

\usepackage[utf8]{inputenc}

\usepackage{microtype}

\usepackage{inconsolata}

\usepackage{graphicx}   %
\usepackage{caption}    %
\usepackage{cuted}
\usepackage{amsmath}
\usepackage{booktabs}
\usepackage[capitalize, noabbrev]{cleveref}

\definecolor{bluehighlight}{RGB}{180, 210, 255}
\definecolor{redhighlight}{RGB}{255, 200, 200}
\definecolor{yellowhighlight}{RGB}{255, 240, 150}
\definecolor{greenhighlight}{RGB}{180, 255, 180}
\definecolor{purplehighlight}{RGB}{230, 220, 255}

\usepackage{pifont}
\newcommand{\cmark}{\textcolor{green}{\ding{51}}} %
\newcommand{\xmark}{\textcolor{red}{\ding{55}}}   %

\title{CoV: Chain-of-View Prompting for Spatial Reasoning}

\author{First Author \\
  Affiliation / Address line 1 \\
  Affiliation / Address line 2 \\
  Affiliation / Address line 3 \\
  \texttt{email@domain} \\\And
  Second Author \\
  Affiliation / Address line 1 \\
  Affiliation / Address line 2 \\
  Affiliation / Address line 3 \\
  \texttt{email@domain} \\}

\author{Haoyu Zhao$^{1*}$\quad
Akide Liu$^{2*}$\quad
Zeyu Zhang$^{2*}$\quad
Weijie Wang$^{1*}$\\
\textbf{Feng Chen}$^{3}$\quad
\textbf{Ruihan Zhu}$^{1}$\quad
\textbf{Gholamreza Haffari}$^{2}$\quad
\textbf{Bohan Zhuang}$^{1\dag}$\\
\vspace{0.2cm}
$^1$ZIP Lab, Zhejiang University\quad
$^2$Monash University\quad
$^3$AIML, Adelaide University\\
\small $^*$Equal contribution. $^\dag$Corresponding author: bohan.zhuang@gmail.com.}

\begin{document}
\maketitle

\begin{strip}
    \vspace{-6em}
    \centering
    \includegraphics[width=\linewidth]{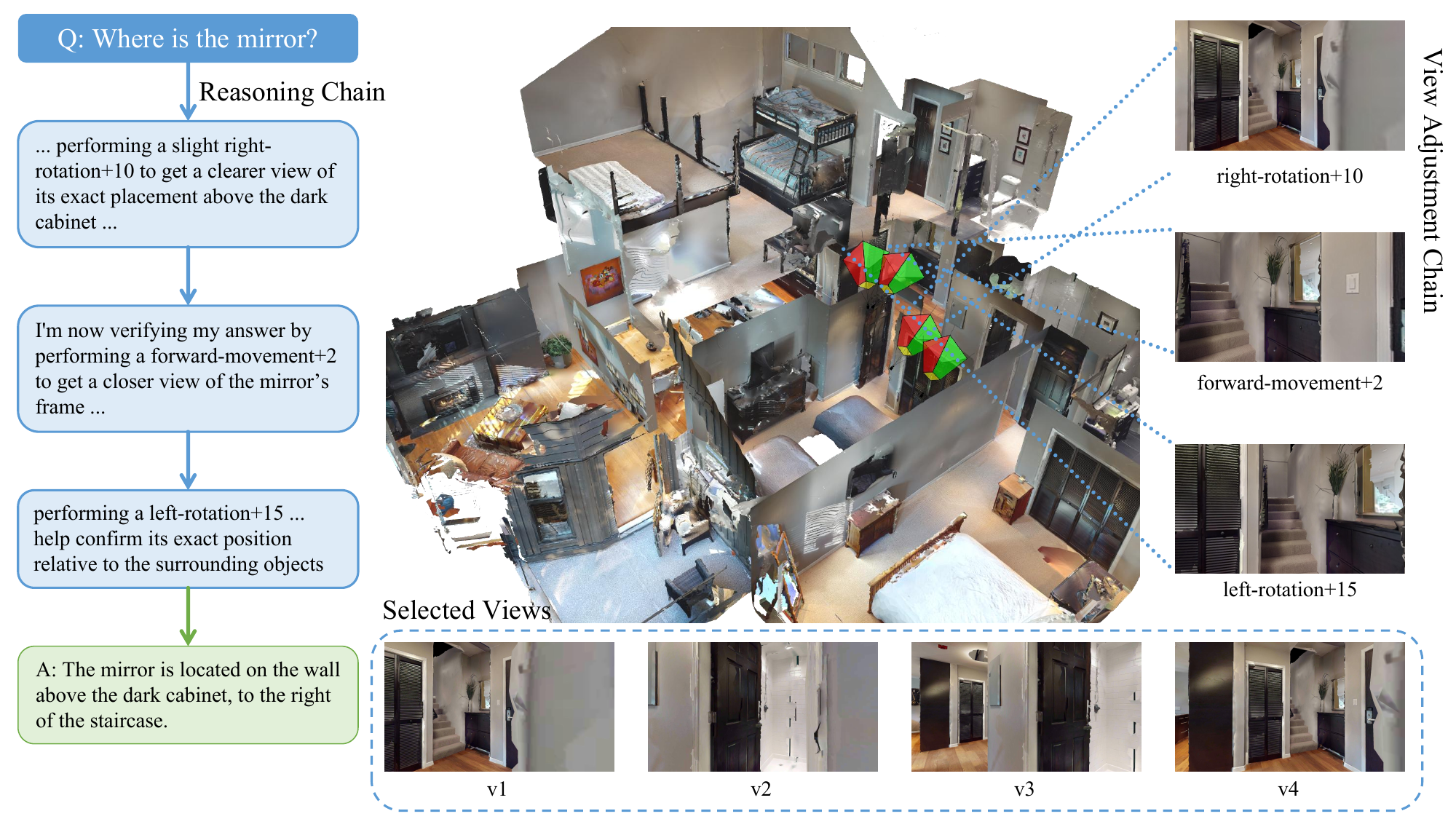}
    \captionof{figure}{The \textbf{Chain-of-View} prompting framework. Given a spatial query and its corresponding 3D scene, CoV facilitates a coarse-to-fine active reasoning process to derive the answer. \textbf{(Bottom)} $v_1$ to $v_4$ denote the task-relevant viewpoints strategically selected from the initial candidate view set by our View Selection Agent. \textbf{(Left and Right)} The interleaved action-reasoning chain demonstrates how the agent dynamically adjusts its perspective (e.g., rotation and movement) to gather discriminative visual evidence and resolve spatial ambiguities. \textbf{(Center)} The visualized camera frustums depict the autonomous exploration trajectory, where the agent bridges the gap between fragmented local views and global spatial context to reach a grounded conclusion.}
    \label{fig:teaser}
    \vspace{-0.5em}
\end{strip}

\begin{abstract}

    Embodied question answering (EQA) in 3D environments often requires collecting context that is distributed across multiple viewpoints and partially occluded. However, most recent vision--language models (VLMs) are constrained to a fixed and finite set of input views, which limits their ability to acquire question-relevant context at inference time and hinders complex spatial reasoning. We propose \textbf{Chain-of-View (CoV) prompting}, a \emph{training-free, test-time} reasoning framework that transforms a VLM into an active viewpoint reasoner through a coarse-to-fine exploration process. CoV first employs a \emph{View Selection} agent to filter redundant frames and identify question-aligned anchor views. It then performs \emph{fine-grained view adjustment} by interleaving iterative reasoning with discrete camera actions, obtaining new observations from the underlying 3D scene representation until sufficient context is gathered or a step budget is reached.
    We evaluate CoV on OpenEQA across four mainstream VLMs and obtain an average +11.56\% improvement in LLM-Match, with a maximum gain of +13.62\% on Qwen3-VL-Flash. CoV further exhibits test-time scaling: increasing the minimum action budget yields an additional +2.51\% average improvement, peaking at +3.73\% on Gemini-2.5-Flash. On ScanQA and SQA3D, CoV delivers strong performance (e.g., 116 CIDEr / 31.9 EM@1 on ScanQA and 51.1 EM@1 on SQA3D). Overall, these results suggest that question-aligned view selection coupled with open-view search is an effective, model-agnostic strategy for improving spatial reasoning in 3D EQA without additional training.
    \texttt{Code:} \url{https://github.com/ziplab/CoV}.
\end{abstract}

\section{Introduction} 

As artificial intelligence transitions from digital domains to physical reality, Embodied Question Answering (EQA) has emerged as a critical capability for enabling intuitive, human-centric interaction with the environment. It holds significant potential across domains like robotics, autonomous navigation, and human–computer interaction. 
In EQA settings, the agent processes a textual question based on a sequence of egocentric images (optionally with a 3D scene representation such as point clouds or 3D meshes).
The agent must perceive and reason within the real environment to derive the correct answer.

\begin{figure}[t]
    \centering
    \includegraphics[width=\linewidth]{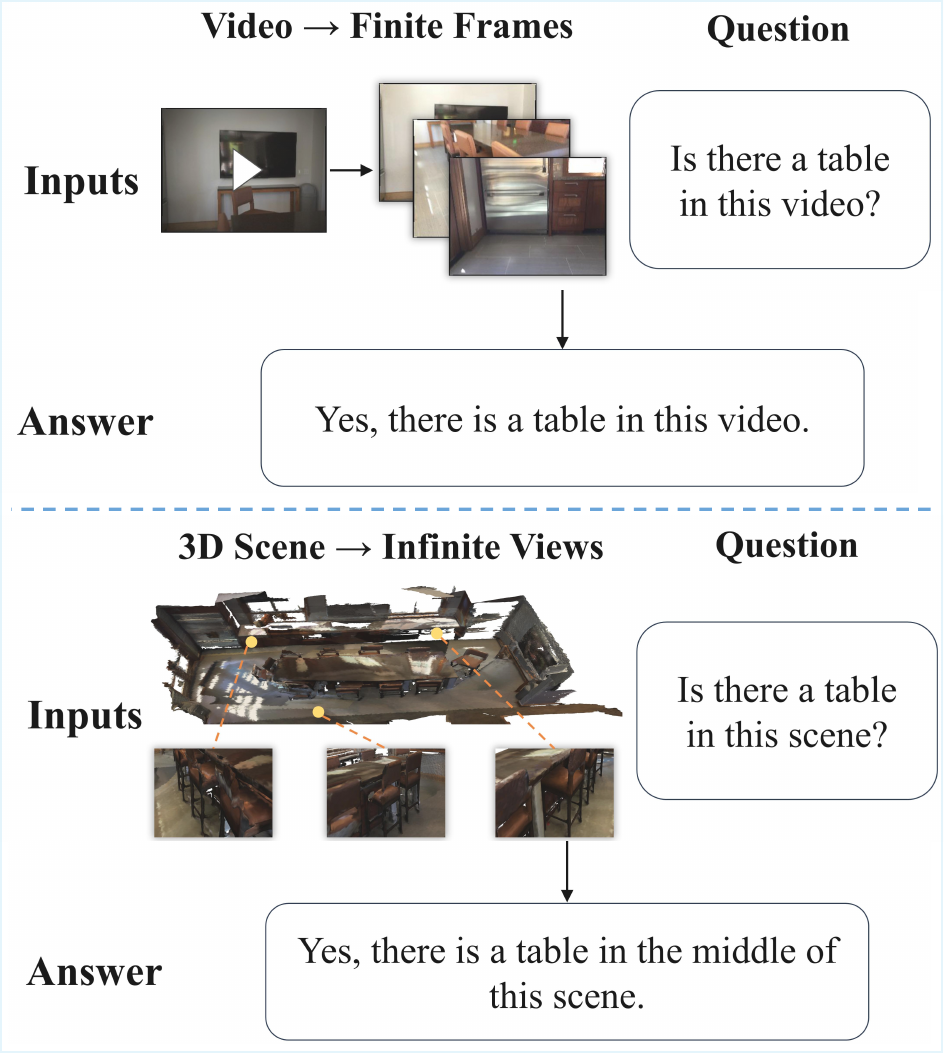}
    \caption{\textbf{Video VLM vs. CoV.} Unlike prior approaches  \textbf{(top)} that rely on fixed-frame video inputs and answer from a limited temporal window, our chain-of-view framework \textbf{(bottom)} explores an open-ended view space constructed from a 3D scene. CoV dynamically selects informative viewpoints and performs step-by-step reasoning during inference, enabling more complete and grounded answers without additional training.}
    \label{fig:vlm_vs_cov}
\end{figure}

However, existing methods~\cite{mo2024bridgeqa, fu2024scenellm, zhu2024llava, li2024video3dllm} encounter a substantial limitation:
conventional methods use a limited and fixed set of viewpoints as input (see~\cref{fig:vlm_vs_cov}), making it difficult for VLMs to acquire sufficient question-relevant views.
In complex embodied QA tasks, answers are not immediately apparent, and a question often requires multi-step reasoning to solve.
For example, for the question “Where can I get some pop drinks?”, the scene does not directly show soda. The model must invoke world knowledge and navigate autonomously to locate objects like a refrigerator. Answering such complex real-world questions requires sufficient question-relevant context and cannot be accomplished through one-step answer generation.

In this paper, we propose the \textbf{chain-of-view (CoV)} prompting framework (see \cref{fig:teaser}), a two-stage agent system designed to shift from passive observation to active exploration and iterative reasoning.
Specifically, the framework operates in a coarse-to-fine manner:
in the coarse-grained view selection stage, the \textbf{View Selection Agent} selects the most question-relevant view from the available viewpoints as the starting point for exploration.
Meanwhile, we provide the agent with a bird-eye view of the entire scene to facilitate a global understanding of the environment.
In the fine-grained view adjustment stage, the \textbf{CoV Agent} executes an action–reasoning loop.
At each step, the VLM generates an action instruction based on the current observation and the question (e.g., forward-movement or right-rotation).
The action is mapped to a rigid-body camera transformation, producing the next observation, which is fed back to the VLM for the next reasoning step.
The process terminates when the CoV agent determines that sufficient information has been acquired or when a predefined limit on action steps is reached, at which point the final answer is produced.

Unlike prior 2D VLMs that rely on predetermined views (as illustrated in~\cref{fig:vlm_vs_cov}), our chain-of-view framework addresses complex embodied QA problems through multi-step reasoning.
Moreover, we improve the alignment between visual content and the question via explicit view selection and fine-grained adjustment.

To evaluate the efficacy of our methodology, we conduct comprehensive experiments on the latest EQA benchmark OpenEQA~\citep{openeqa}, which serve as standard metrics for assessing 3D scene understanding and question answering capabilities. 
We evaluate both mainstream open-source and proprietary VLMs, and applying the CoV framework yields an average improvement of $10.82\%$, with a maximum gain of $13.62\%$ on Qwen3-VL-Flash.
We further empirically verify the test-time scaling capability of CoV: as the number of action steps increases, the agent’s score improves by an average of $2.51\%$, with a maximum gain of $3.73\%$ on Gemini-2.5-Flash.

Qualitative analyses further validate that our CoV framework produces more coherent and interpretable reasoning chains, particularly in complex or cluttered environments. 
These results demonstrate the potential of test-time scaling strategies to enhance scene understanding without requiring additional model training or dataset-specific tuning, making our framework robust and adaptable across diverse 3D tasks and domains.

The main contributions of our work can be summarized as follows:
\begin{itemize}
    \item We propose Chain-of-View Prompting, a test-time reasoning framework that enhances VLMs’ ability to handle complex spatial reasoning in embodied question answering. By leveraging coarse-to-fine view selection and camera adjustment, the agent acquires sufficient question-relevant views to answer spatially complex questions, thereby improving performance on EQA tasks.
    
    \item CoV prompting enables test-time scaling. As the number of exploration steps increases, the agent’s performance improves gradually.
    
    \item Experimental results on the latest embodied QA benchmarks demonstrate significant improvements through our systematic view exploration approach. Our method achieves up to a $13.62\%$ improvement on the OpenEQA benchmark.
\end{itemize}

\section{Related Work}

Recent advances in 3D scene understanding have unified perception and language.
Methods like Vote2Cap-DETR~\cite{vote2cap2023}, D3Net~\cite{chen2021d3net}, and SpaCap3D~\cite{spa2cap2022} integrate object localization and description generation, enabling more grounded scene understanding for robotics, AR/VR, and embodied AI.
3D vision-language models such as LLaVA-3D~\cite{zhu2024llava} and LL3DA~\cite{ll3da} further advance scene understanding by synthesizing 2D multimodal perception with 3D spatial context. 
These architectures leverage multi-view images augmented with 3D positional embeddings, facilitating more context-aware reasoning without dependency on external object proposals or segmentation mechanisms.

\paragraph{Test-time reasoning.}

Large models such as Qwen, ChatGPT and Gemini \cite{bai2025qwen3vltechnicalreport,openai2023gpt4,georgiev2024gemini}
show strong performance in multi-modal reasoning tasks. 
Due to high fine-tuning costs, recent work explores efficient adaptation methods that keep pretrained weights. 
In-context learning~\cite{brown2020language,sahoo2024systematic}, prompt engineering, and chain-of-thought prompting~\cite{cot} guide model behavior at inference time.
Recent works like Simple Scaling~\cite{muennighoff2025simple}, adaptive compute~\cite{snell2024scaling}, and calibration~\cite{mckenna2024calibrating} offer practical, training-free improvements.

\paragraph{Scene understanding.}

3D scene understanding primarily encompasses tasks such as 3D question answering \cite{ma2023sqa3d,scanqa} and 3D dense captioning \cite{chen2020scanrefer,achlioptas2020referit_3d}. 
Early 3D dense captioning used a detect-then-describe pipeline~\cite{scan2cap,spa2cap2022}. 
Newer methods adopt end-to-end transformers \cite{vote2cap2023,chen2021d3net,huang20253d,huang2025dc} to predicts object-caption pairs directly. 
Methods for embodied QA often incorporate multi-modal fusion~\cite{mo2024bridgeqa}, navigation-conditioned reasoning~\cite{zheng2024navillm}, and LLM grounding in 3D scenes~\cite{3dllm} to improve spatial and semantic understanding.

\paragraph{3D VLMs.}

Recent 3D VLM works integrate point clouds~\cite{huang2023chat,ll3da,yang2023leo,zhang2024chatscene} and multi-view images~\cite{fu2024scenellm,3dllm,qi2025gpt4scene} into large language models for scene reasoning. LL3DA~\cite{ll3da} encodes global features from scene-level point clouds. LEO~\cite{yang2023leo} and Chat-Scene~\cite{zhang2024chatscene} segment and encode object-level features. 
3D-LLM~\cite{3dllm} and Scene-LLM~\cite{fu2024scenellm} use object-centric patches from multi-view images. 
LLaVA-3D~\cite{zhu2024llava} builds on 2D LMMs with 3D positional embeddings to structure image patches spatially.

\section{Method}
\label{method}

\begin{figure*}[t]
  \centering
  \includegraphics[width=\linewidth]{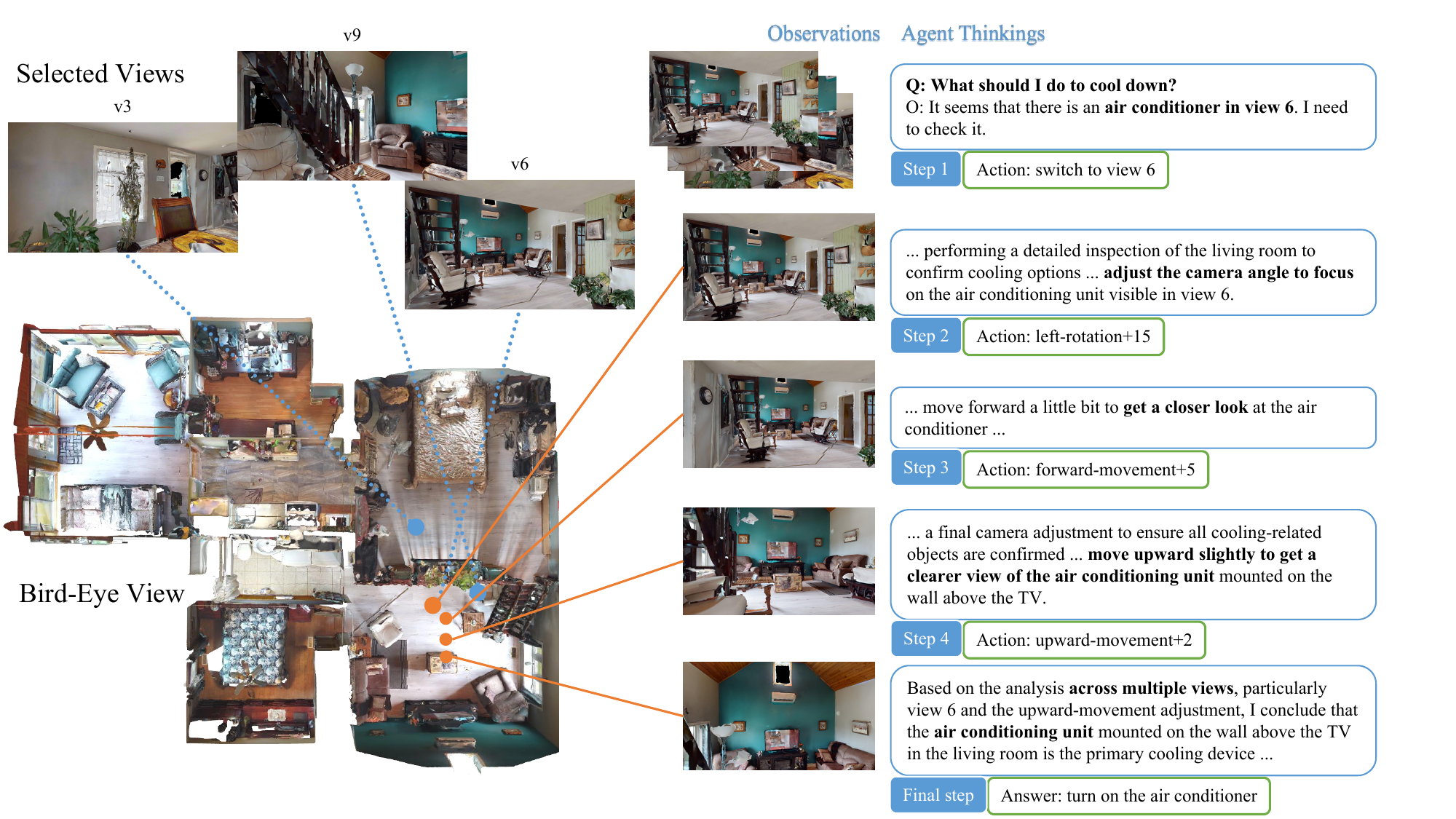}
  \caption{\textbf{Action-reasoning chain of the CoV agent.} The CoV agent executes an iterative action–reasoning chain. For the question “What should I do to cool down?”, the agent first selects view 6 from the input images as an anchor. It then adjusts the viewpoint at each reasoning step to acquire new observations. Once the agent determines that sufficient information has been obtained, it outputs the answer “turn on the air conditioner.”}
  \label{fig:cov-agent}
\end{figure*}

\subsection{Problem Setting}

The goal of the embodied question answering task is to answer questions related to a 3D scene.
The questions can span multiple categories, such as object recognition, attribute recognition, object localization, and spatial reasoning etc.~\citep{openeqa}. 
Unlike conventional 2D VQA with fixed frame sequences, embodied question answering utilizes either multi-view RGB-D images with camera poses or reconstructed 3D meshes, enabling reasoning from arbitrary viewpoints aligned with question semantics.

Formally, let $S$ denote the 3D scene representation (e.g., a point cloud or mesh) and $Q$ be the natural language query. A 3D scene implies a continuous space of potential viewpoints $\Omega$. 
However, directly reasoning over the raw global geometry $S$ is often inefficient and lacks fine-grained visual details.
In practice, the input for embodied question answering is combined with a sequence of frames $\mathcal{V} = \{v_1, \dots, v_T\}$ sampled from a video episode of the scene.
The objective is to generate a textual answer $A$ that accurately addresses the query $Q$ based on these observations.
This process is modeled as maximizing the conditional probability:
\begin{equation}
    A^* = \operatorname*{argmax}_{A} P(A \mid S, \mathcal{V}, Q).
\end{equation}

\subsection{Method Overview}

We propose chain-of-view prompting, a test-time search framework for embodied visual question answering.
The goal of CoV prompting is to infer the correct answer $A$ from a given question text $Q$ and the corresponding 3D scene.
The overall pipeline consists of two primary stages: (i) Coarse-grained View Selection, and (ii) Fine-grained View Adjustment. 
Initially, we filter the full set of input view frames to a subset of candidate views based on their relevance to the question. 
We dynamically adjust the perspectives within this subset to produce a refined sequence of viewpoints. 
Finally, an answer is synthesized based on these optimized viewpoints.

\subsection{Coarse-Grained View Selection}

$\mathcal{V}$ is sampled from the video episode of the corresponding scene, containing many redundant frames with low information density.
Only a small number of key frames are relevant to the question $Q$, while a large amount of irrelevant information can interfere with the agent’s correct judgment.
Therefore, we employ a View Selection Agent to filter the initially available views and extract the frames most relevant to the question.
The prompt template of the view selection agent is provided in~\cref{appendix:prmompts}.
Given the input $(Q, \mathcal{V})$, the view selection agent selects a reduced subset of frames $\mathcal{V}' = \{v_{i_1}, \dots, v_{i_K}\}$, where $K \ll T$.

This initial filtering step significantly reduces redundancy in the input data and focuses on views that are most likely to contain information relevant to answering the question. 
By narrowing down the search space, we enable more efficient processing in the subsequent fine-grained phase, effectively addressing the challenge of searching through the vast view space of 3D scenes.

\subsection{Fine-Grained View Adjustment} 

In prior EQA approaches, the agent passively infers an answer over a limited and fixed set of input frames and cannot actively acquire informative observations from the environment. This one-step generation paradigm overlooks environmental details that may be relevant to the question and constrains the agent’s performance.
Inspired by chain-of-thoughts~\citep{cot}, we aim to provide more detailed environmental information through fine-grained viewpoint adjustment while eliciting deeper thinking.
Therefore, we employ the Chain-of-View Agent for fine-grained view adjustment. The prompt template is provided in \cref{appendix:prmompts}.

We leverage the 3D scene representation to dynamically generate new perspectives that reveal information not visible from the initially selected views. 
For each initial view $v_i \in \mathcal{V}'$, the CoV agent generates a sequence of actions $\{a_1, \dots, a_L\}$, where $a_t \in \mathcal{A}$ for $t = 1, \dots, L$, and $\mathcal{A}$ denotes the agent's action space. It consists of discrete translational and rotational actions:
\begin{itemize}
    \item \textbf{Translational actions:} move forward, move backward, move left, move right, move up, and move down. Each action corresponds to a fixed displacement along the respective axis of the agent's local coordinate frame.
    \item \textbf{Rotational actions:} yaw (rotate left/right), pitch (look up/down), and roll (tilt clockwise/counterclockwise). Each rotation adjusts the agent's orientation around its local coordinate axes.
    \item \textbf{View switch actions:} switch to $v_i, v_i \in \mathcal{V}'$. Agent can switch to any view anchor obtained through the coarse view selection stage.
\end{itemize}

Specifically, the agent’s context is defined as $C_0 = \{Q, \mathcal{V}'\}$ at start.
At each step, the CoV agent thinks over the current observation and the question to generate an action instruction $a_t$.
We convert $a_t$ into an $\mathrm{SE}(3)$ transformation matrix, which updates the camera pose and yields a new viewpoint $v_i^{t+1}$:
\begin{equation}
{v}_i^{t+1} = \text{Transform}(v_i; a_t, S).
\end{equation}
$v_i^{t+1}$ is appended into the agent’s context,which is then fed into the model for the next step:
\begin{equation}
    C_{t+1} = \{Q, V', v_i^{1}, \ldots, v_i^{t+1}\}.
\end{equation}
The reasoning process terminates when the CoV agent determines that sufficient information has been collected to answer the question or when a predefined action step limit is reached, at which point the final answer is produced.

Figure~\ref{fig:cov-agent} presents a multi-step reasoning example of the CoV agent. 
Fine-grained view adjustment enables the agent to observe regions that were previously occluded or blurred, allowing it to acquire richer question-relevant environmental details and thereby answer the question more accurately.

\section{Experiments}
\label{exp}

\begin{figure*}[t]
  \centering
  \includegraphics[width=\linewidth]{figures/Figure4.png}
  \caption{Visualization of CoV reasoning results. Our method selects informative views and produces coherent multi-step answers grounded in the spatial context.}
  \label{fig:vis}
\end{figure*}

\subsection{Benchmarks and Metrics}
We evaluate our method on the OpenEQA~\citep{openeqa}, ScanQA~\citep{scanqa}, and SQA3D~\citep{ma2023sqa3d} benchmarks, covering both mainstream open-source and proprietary models.

Sourced from over 180 real-world environments (ScanNet and HM3D~\citep{dai2017scannet, ramakrishnan2021hm3d}), OpenEQA is a challenging open-vocabulary benchmark, designed to evaluate embodied question answering capabilities in the era of foundation models. 
Besides, we include ScanQA and SQA3D, two representative datasets for spatial reasoning challenges. 
ScanQA is a large-scale 3D question answering dataset comprising over 41,000 question-answer pairs. It focuses on object-grounded QA by linking natural language queries to specific 3D objects within richly annotated RGB-D scans, facilitating spatial reasoning and object localization. In contrast, SQA3D emphasizes situated reasoning, requiring agents to understand their position and orientation within a 3D scene. It includes 33,400 reasoning questions spanning 6,800 unique situations from 650 ScanNet scenes, presenting complex challenges such as spatial relationships, commonsense understanding, and multi-hop inference.

We use the \textbf{LLM-Match} metric proposed in OpenEQA.
This metric utilizes an LLM judge to compare a predicted answer against the ground-truth answer, assigning a score $\gamma_i \in \{1, \dots, 5\}$. 
The final metric is computed by normalizing and averaging them to a percentage scale over question number $N$:
\begin{equation}
  \text{LLM-Match} = \frac{1}{N} \sum_{i=1}^{N} \left( \frac{\gamma_i - 1}{4} \right) \times 100\%.
\end{equation}

For ScanQA and SQA3D, we adopt a comprehensive set to assess answer accuracy. Specifically, CIDEr (C) measures consensus with human annotations; BLEU-4 (B-4) captures n-gram overlap; METEOR (M) considers both precision and recall with synonym matching; ROUGE-L (R) evaluates the longest common subsequence; and Exact Match at top-1 (EM\@1) reflects the strict correctness of generated answers. 

\subsection{Implementation Details}
The input for each question–answer pair consists of video frames uniformly sampled from a scene video episode at a ratio of 10:1, together with the textual question $Q$.
For the baseline, we provide all images in a single pass and let the model directly generate an answer.
For CoV, we first feed all images to the View Selection agent for coarse filtering, and the selected views are then passed to the CoV agent to answer the question; both agents use the same underlying VLM.
The prompt templates used are provided in~\cref{appendix:prmompts}.

\begin{table*}[t]
\centering
\caption{\textbf{Quantitative comparison with SOTA models on ScanQA (val) and SQA3D (test)}. ``C'' stands for ``CIDEr'', ``B-4'' for ``BLEU-4'', ``M'' for ``METEOR'', ``R'' for ``ROUGE'', and ``EM@1'' for top-1 exact match. BEV denotes whether a model takes bird-eye view as an input.}
\label{tab:scanqa_sqa3d_metrics}

\resizebox{\linewidth}{!}{

\begin{tabular}{lccccccccc}
    \toprule
     & & & \multicolumn{6}{c}{ScanQA (val)} & \multicolumn{1}{c}{SQA3D (test)} \\
     \cmidrule(lr){4-9} \cmidrule(lr){10-10}
     
    Model & Venue & BEV & C & B-4 & M & R & EM@1 & & EM@1 \\
      
    \midrule
    \rowcolor{bluehighlight} \multicolumn{10}{l}{\textbf{Task-specific models}} \\
    
    Scan2Cap~\cite{scan2cap} & - & \xmark & - & - & - & - & - & & $41.0$ \\
    ScanRefer+MCAN~\cite{scanrefer_mcan} & ECCV2020 & \xmark & $55.4$ & $7.9$ & $11.5$ & $30.0$ & $18.6$ & & - \\
    ClipBERT~\cite{clipbert} & CVPR2021 & \xmark & - & - & - & - & - & & $43.3$ \\
    ScanQA~\cite{scanqa} & CVPR2022 & \xmark & $64.9$ & $10.1$ & $13.1$ & $33.3$ & $21.1$ & & $47.2$ \\
    3D-VisTA~\cite{3d-vista} & ICCV2023 & \xmark & $69.6$ & $10.4$ & $13.9$ & $35.7$ & $22.4$ & & $48.5$ \\
    
    \midrule
    \rowcolor{redhighlight} \multicolumn{10}{l}{\textbf{3D LMMs}} \\
    3D-LLM (FlanT5)~\cite{3dllm} & NeurIPS2023 & \xmark & $69.4$ & $12.0$ & $14.5$ & $35.7$ & $20.5$ & & - \\
    LL3DA~\cite{ll3da} & CVPR2024 & \xmark & $76.8$ & $13.5$ & $15.9$ & $37.3$ & - & & - \\
    Chat-3D v2~\cite{chat3d-v2} & - & \xmark & $87.6$ & $14.0$ & - & - & - & & $54.7$ \\
    LEO~\cite{leo} & ICML2024 & \xmark &$101.4$ & $13.2$ & $20.0$ & $49.2$ & $24.5$ & & $50.0$ \\
    Scene-LLM ~\cite{fu2024scenellm} & WACV2025 & \xmark & $80$ & $12.0$ & $16.6$ & $40.0$ &$27.2$ & & $54.2$ \\
    ChatScene ~\cite{chatscene} & CVPR2024 & \xmark & $87.7$ & $14.3$ & $18.0$ & $41.6$ & $21.6$ & & $54.6$ \\
    
    \midrule
    \rowcolor{yellowhighlight} \multicolumn{10}{l}{\textbf{Zero-shot 2D LMMs}} \\
    VideoChat2~\cite{videochatv2} & CVPR2024 & \cmark & $49.2$ & $9.6$ & $9.5$ & $28.2$ & $19.2$ & & $37.3$ \\
    LLaVA-NeXT-Video~\cite{llava-onevision} & - & \cmark & $46.2$ & $9.8$ & $9.1$ & $27.8$ & $18.7$ & & $34.2$ \\
    LLaVA-Video~\cite{llava-video} & - & \cmark & $88.7$ & - & - & - & - & & $48.5$ \\
    
    \midrule

    \rowcolor{purplehighlight} CoV (Ours) & - & \cmark & $\mathbf{116}$ & 
    $\mathbf{16.9}$ & $\mathbf{24.5}$ & $\mathbf{50.4}$ & $\mathbf{31.9}$ & & $51.1$ \\
    \bottomrule
\end{tabular}

}

\end{table*}

\subsection{Main Results}

For ScanQA and SQA3D, we use GPT-4.1, Gemini Pro Flash, and InternVL for comparison~\citep{openai2023gpt4,georgiev2024gemini,chen2023internvl}. 
We run all evaluations in the val-unseen set for fair, zero-shot comparison with 3D VLM baselines.
\Cref{tab:scanqa_sqa3d_metrics} presents that our method achieves state-of-the-art performance with notable improvements in CIDEr (116 vs. LEO's 101.4) and EM@1 (31.9\% on ScanQA). 
These results demonstrate our method's effectiveness in both generating human-like responses and providing accurate answers.

For OpenEQA~\citep{openeqa}, we evaluate Qwen3-VL-Flash, GLM-4.6V, Gemini-2.5-Flash and GPT-4o-mini~\citep{bai2025qwen3vltechnicalreport, vteam2025glm45vglm41vthinkingversatilemultimodal, comanici2025gemini25pushingfrontier, openai2024gpt4technicalreport}.
The evaluation results are shown in~\cref{tab:openeqa}.
Compared to the setting without CoV, our method achieves an average improvement of $11.56\%$ and a maximum gain of $13.62\%$ (on Qwen3-VL-Flash) in a training-free manner.

\begin{table*}[t]
\centering
\caption{
\textbf{Performance comparison on the OpenEQA benchmark.} During evaluation, we fix the temperature of each model to 0. 
LLM-Match is measured using the evaluation protocol provided by OpenEQA.
$n$ step denotes that the minimum number of action steps for the model is set to $n$.
\textbf{Scaling Improvement} measures the gain obtained by increasing the number of action steps, defined as the difference between the best CoV result and CoV-1.
\textbf{Boldface} indicates the \textbf{best} results.
}
\label{tab:openeqa}

\begin{tabular}{l c c c c}
    \toprule
   Method & Qwen3-VL-Flash & GLM-4.6v & GPT-4o-Mini & Gemini-2.5-Flash \\
    \midrule
    Baseline & $52.65$ & $62.40$ & $45.87$ & $52.30$ \\
    \midrule
    CoV(1 Step) & $58.75$ & $67.00$ & $49.85$ & $57.10$ \\
    2 Step & $59.07$ & $67.33$ & $50.15$ & $57.15$ \\
    3 Step & $59.03$ & $67.20$ & $49.60$ & $58.80$ \\
    4 Step & $\mathbf{59.82}$ & $66.30$ & $50.80$ & $58.15$ \\
    5 Step & $59.03$ & $67.23$ & $\mathbf{51.56}$ & $58.82$ \\
    6 Step & $59.39$ & $\mathbf{67.70}$ & $51.41$ & $\mathbf{59.23}$ \\
    7 Step & $59.77$ & $67.64$ & $51.62$ & $59.16$ \\
    \midrule
    \textbf{Improvement} & $\uparrow \mathbf{13.62\%}$ & $\uparrow 8.50\%$ & $\uparrow 12.40\%$ & $\uparrow 11.70\%$ \\
    \textbf{Scaling Improvement} & $\uparrow 1.82\%$ & $\uparrow 1.04\%$ & $\uparrow 3.43\%$ & $\uparrow \mathbf{3.73\%}$ \\
    \bottomrule
\end{tabular}

\end{table*}

Evaluation results confirm that our chain-of-view framework produces higher quality answers by iteratively refining visual understanding through strategically selected views, enabling deep spatial reasoning in complex 3D environments.

\subsection{Test-Time Scaling}

\begin{figure}[t]
    \centering
    \includegraphics[width=\linewidth]{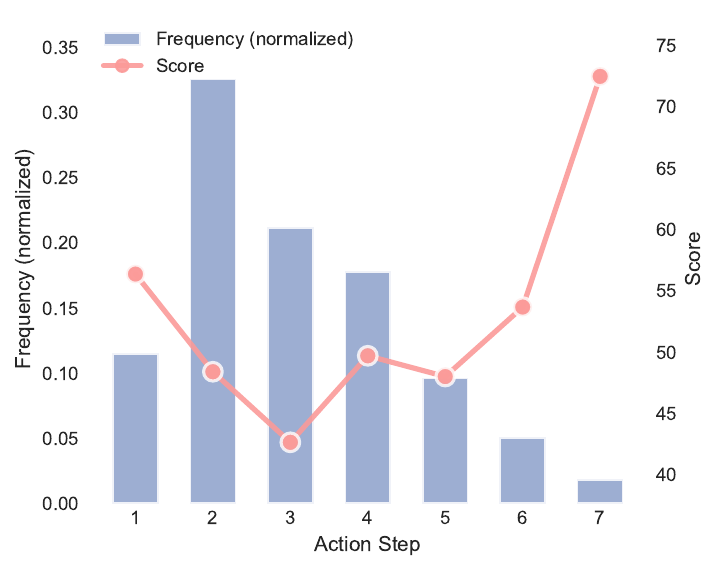}
    \caption{\textbf{Action Step Analysis.} Distribution of action steps for Qwen3-VL-Flash~\citep{bai2025qwen3vltechnicalreport} on the OpenEQA~\citep{openeqa} dataset.}
    \label{fig:step-dist}
\end{figure}

We investigate the test-time scaling behavior of chain-of-view prompting by quantitatively analyzing the relationship between the number of action steps and performance.
When the number of action steps is not constrained, the statistics of question counts and scores at different action steps for the CoV agent on the OpenEQA dataset are shown in~\cref{fig:step-dist}.
Questions that require more action steps exhibit a clear upward trend in scores.
However, most questions involve only a small number of action steps, typically between one and three.
If the agent were able to execute more action steps, its performance on embodied QA tasks would be expected to further improve.

Inspired by the budget forcing strategy used in S1~\citep{muennighoff2025s1}, we set a lower bound on the number of action steps in the CoV agent’s prompt template (see~\cref{appendix:prmompts}).
As shown in~\cref{fig:scaling}, increasing the number of action steps gradually improves.
Compared to setting the minimum number of action steps to 1, increasing the action-step limit yields an average improvement of $2.51\%$.

These results demonstrate that multi-step reasoning enhances agent performance on EQA tasks and highlight the potential of our method for training-free test-time scaling.

\begin{figure}[t]
    \centering
    \includegraphics[width=\linewidth]{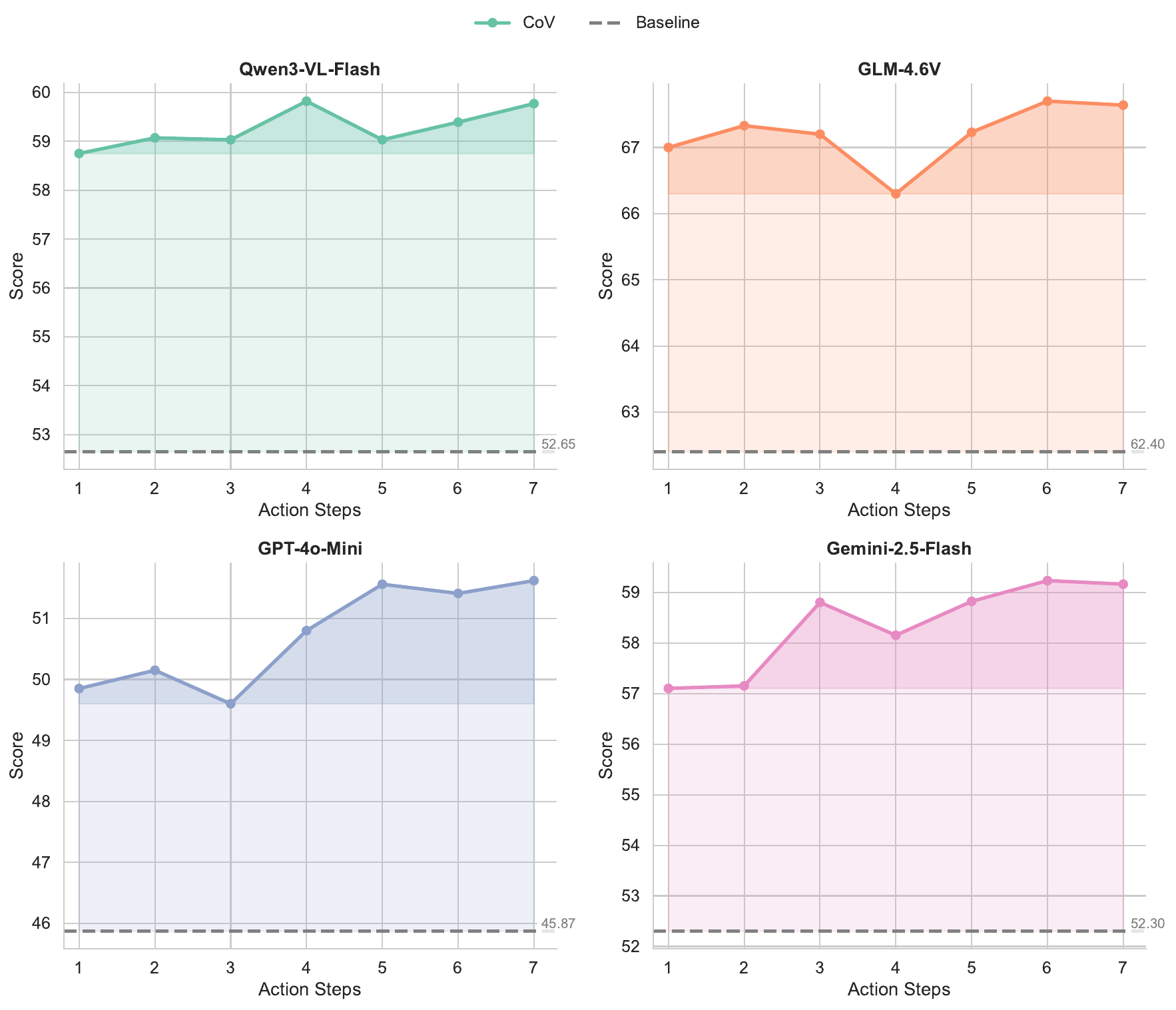}
    \caption{\textbf{Test-time scaling ability of CoV.} On OpenEQA, we evaluate different VLMs and observe that the performance of all models gradually improves as the number of action steps increases.}
    \label{fig:scaling}
\end{figure}
\subsection{Ablation Studies}
We conduct an ablation study to examine the role of coarse-grained view selection.
We evaluate the agent’s performance on OpenEQA~\citep{openai2023gpt4} with and without the coarse-grained view selection agent.
Without the coarse-grained stage to filter views, the CoV agent is exposed to a large number of redundant and low-information-density frames, making it harder to identify a question-relevant anchor to initiate actions.

As shown in~\cref{tab:ablation-view-selection}, removing the view selection agent from our method degrades VLM performance, with an average drop of $4.59\%$.
This result demonstrates that coarse-grained view selection is an essential component of our approach.

\begin{table}[t]
\centering
\caption{
\textbf{Ablation study on view selection component.} This table shows the performance comparison between the baseline, CoV 3, and CoV 3 without the coarse view selection mechanism across different models. CVS denotes coarse view selection.
}
\label{tab:ablation-view-selection}
\resizebox{\linewidth}{!}{

\begin{tabular}{l c c}
    \toprule
    Model & CoV(3 Step) & CoV w/o CVS \\
    \midrule
    Qwen3-VL-Flash & $\mathbf{59.03}$ & $57.50$  \\
    GLM-4.6v & $\mathbf{67.20}$ & $62.43$ \\
    GPT-4o-Mini & $\mathbf{49.60}$ & $46.74$ \\
    Gemini-2.5-Flash & $\mathbf{58.80}$ & $57.11$ \\
    \bottomrule
\end{tabular}

}

\end{table}

\subsection{Qualitative Result}
Figure \ref{fig:vis} presents a qualitative example illustrating how CoV selects and reasons over a sequence of informative views. 
The agent progressively explores the scene, identifying relevant object locations with high spatial precision and integrating these observations into its reasoning process. 
Through multi-step reasoning, CoV generates answers that are not only semantically aligned with the question but also consistent with the spatial layout of the scene. 
The predicted answer closely matches the ground truth, highlighting the agent’s capability to perform detailed spatial reasoning.

Both qualitative and quantitative analyses demonstrate the effectiveness of the CoV method.
Through coarse-grained view selection and multi-step actions, the agent autonomously acquires more question-relevant information and is able to perform complex embodied QA tasks.
Additional qualitative visualization examples are provided in~\cref{appendix:vis-result}.

\section{Conclusion}
\label{sec:conclusion}

In conclusion, our work rethinks embodied question answering through the lens of viewpoint-aware reasoning. 
By adopting a coarse-to-fine viewpoint adjustment strategy, the view selection agent and the CoV agent can acquire question-relevant observations and address complex spatial reasoning problems through multi-step reasoning.
We further examine the test-time scaling capability of the proposed method and observe that the agent’s performance improves as the number of action steps increases.
Beyond performance gains, our framework represents a conceptual shift—emphasizing not just what the model sees, but how it sees, remembers, and reasons. 
We believe this chain-of-view prompting framework will open new possibilities for embodied AI systems that must act, adapt, and explore in complex real-world spaces.

\section*{Limitations}
While the CoV Prompting approach offers significant advancements in embodied question answering, it is not without limitations. 
The coarse-to-fine paradigm may struggle in highly dynamic or cluttered environments where rapid context shifts occur, potentially leading to misinterpretation of scene elements. When action trajectory is too long, excessive exploration may also introduce noise or hallucination.
Additionally, the effectiveness of CoV relies on the quality and relevance of the selected views; suboptimal view selection can impair reasoning accuracy. 
Developing advanced view selection algorithms and adaptive reasoning mechanisms to address these challenges represents a promising direction for future research.

\bibliography{custom}
\clearpage
\appendix
\section{Prompt Templates}
\label{appendix:prmompts}
\begin{figure}[t]
    \centering
    \includegraphics[width=\linewidth]{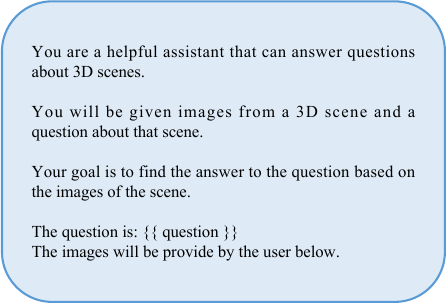}
    \caption{Prompt template for baseline.}
    \label{fig:prompt-baseline}
\end{figure}

\begin{figure}[t]
    \centering
    \includegraphics[width=\linewidth]{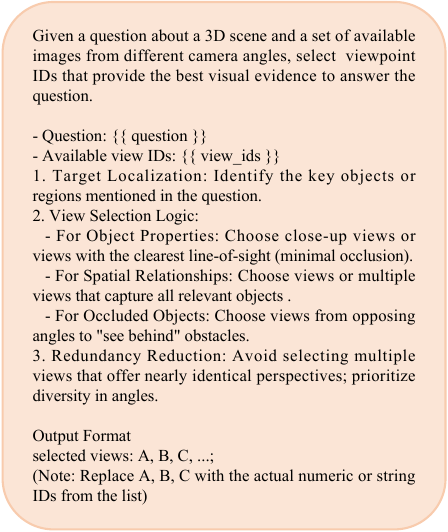}
    \caption{Prompt template for coarse-grained view selection agent.}
    \label{fig:prompt-cvs}
\end{figure}

\begin{figure}[t]
    \centering
    \includegraphics[width=\linewidth]{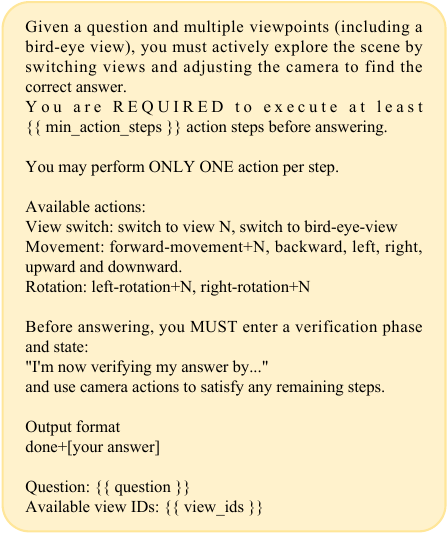}
    \caption{Prompt template for fine-grained CoV agent.}
    \label{fig:prompt-cov}
\end{figure}

The prompt templates used in our experiments are shown in~\cref{fig:prompt-baseline,fig:prompt-cvs,fig:prompt-cov}.
For the baseline, we directly provide all view frames along with the question and let the model generate an answer.
For the view selection agent, we input all view frames, the question, and an index for each frame, and the model is required to output the IDs of the views relevant to the question.
For the CoV agent, we input the coarsely selected views and the question, and the model produces the answer after multi-step action execution.

\section{Result Visualization}
\label{appendix:vis-result}
\begin{figure*}[t]
  \centering
  \includegraphics[page=1, width=\linewidth]{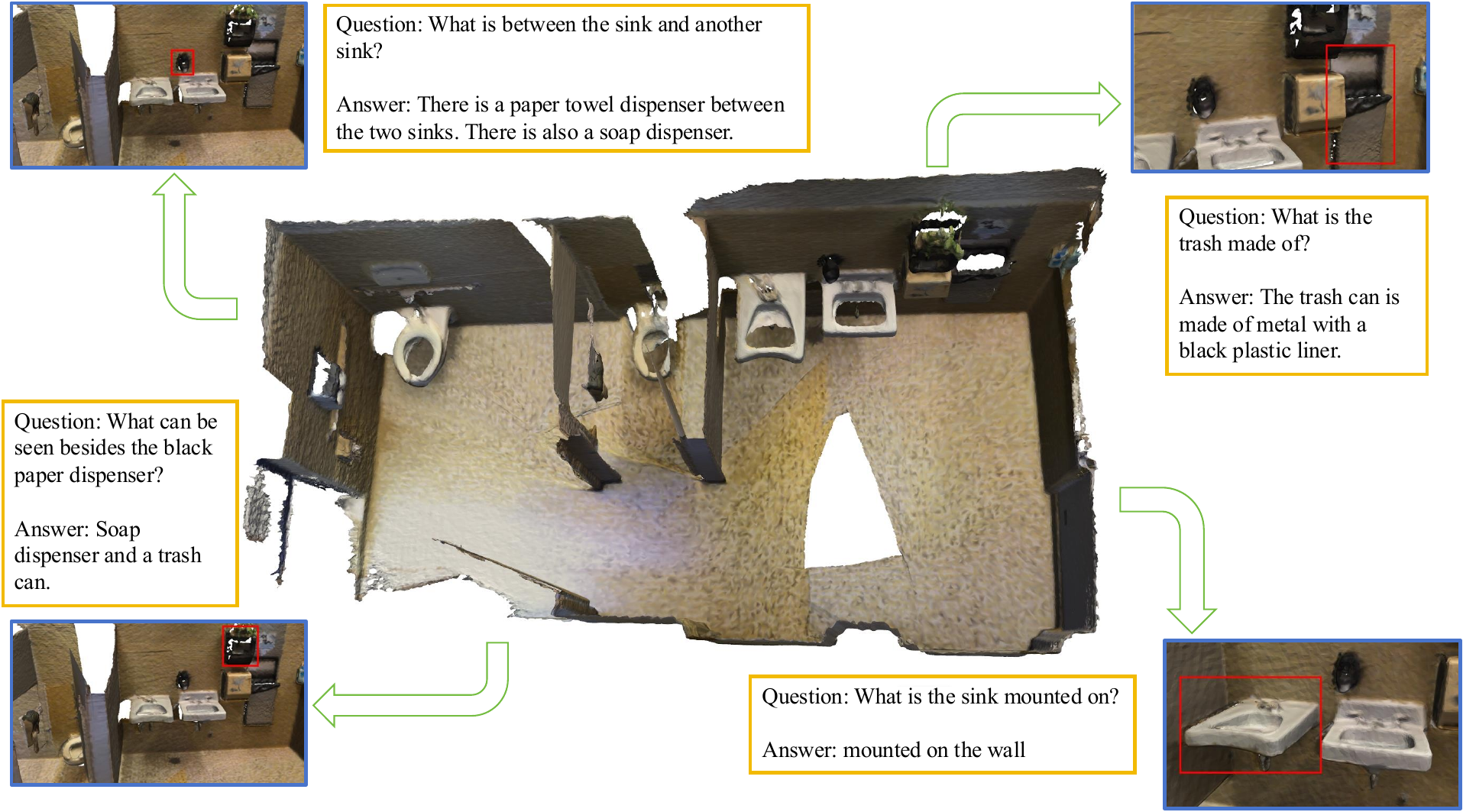}
  \caption{CoV identifies the soap and paper towel dispensers between two sinks and reasons about material and spatial attributes such as the metal trash can and wall-mounted sink.}
  \label{fig:pdf_page1}
\end{figure*}

\begin{figure*}[t]
  \centering
  \includegraphics[page=2, width=\linewidth]{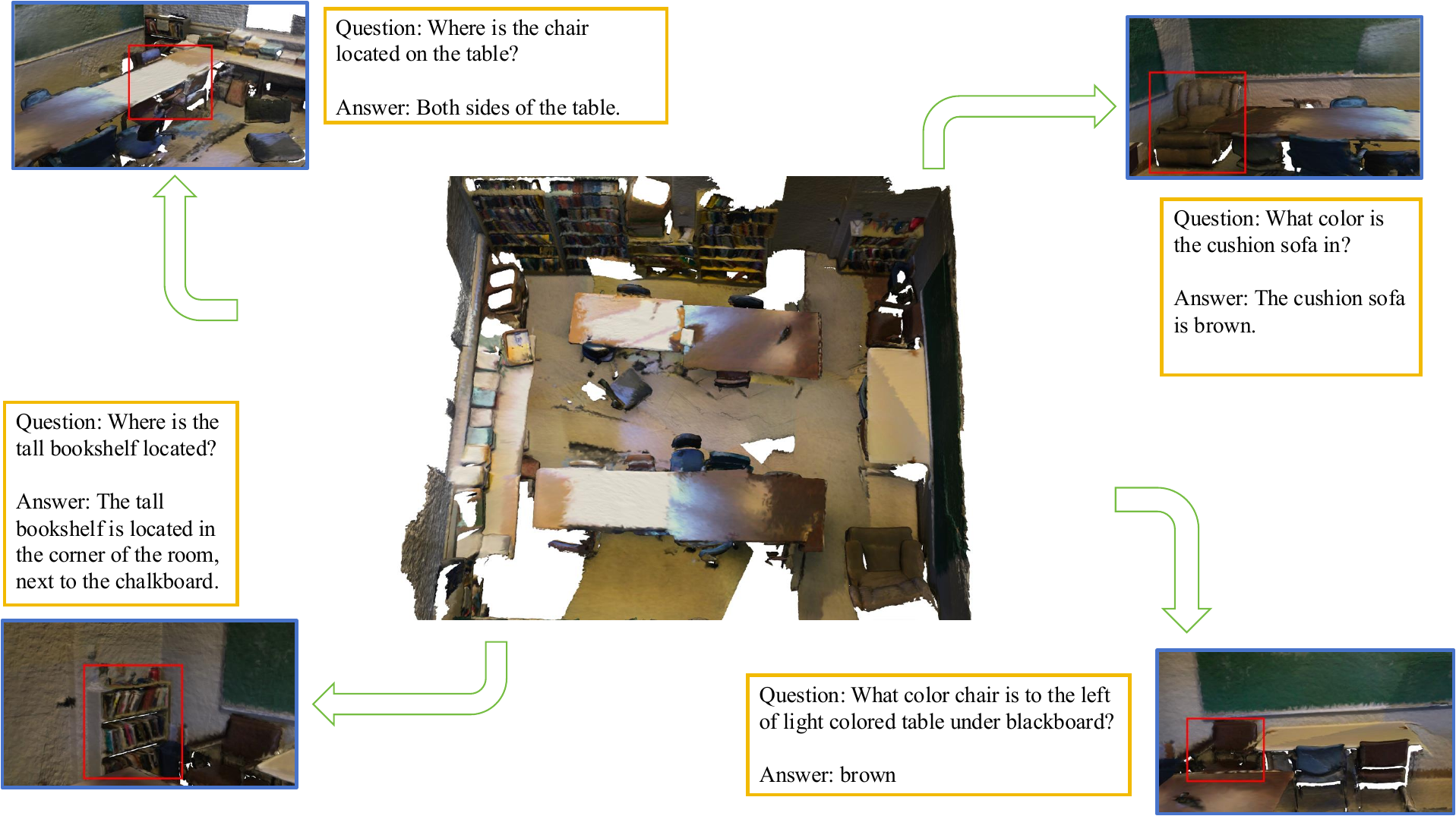}
  \caption{CoV accurately localizes furniture such as brown cushion sofas, chairs beside a table, and a tall bookshelf positioned near the chalkboard, demonstrating spatial alignment and object identification.}
  \label{fig:pdf_page2}
\end{figure*}

\begin{figure*}[t]
  \centering
  \includegraphics[page=3, width=\linewidth]{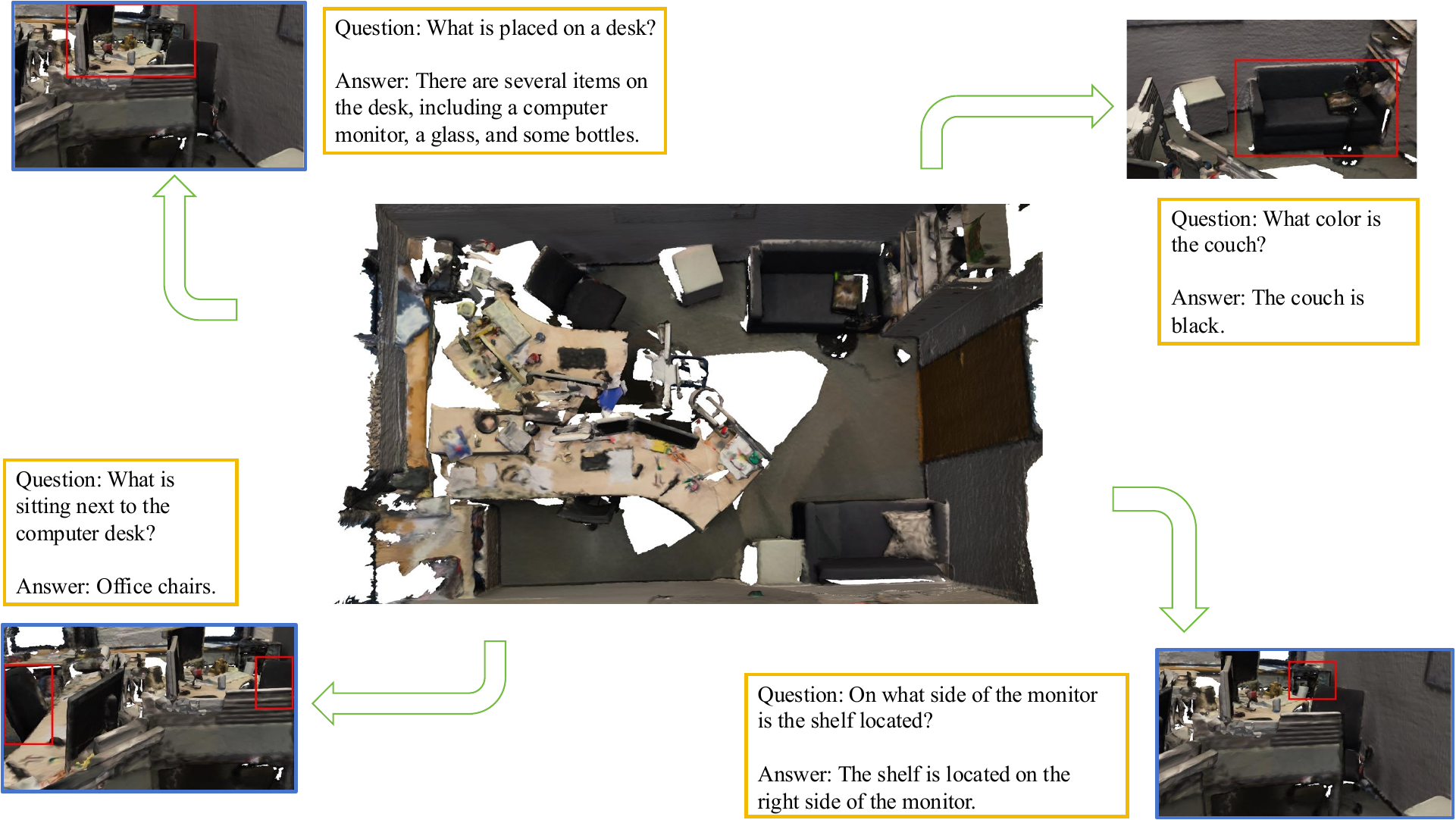}
  \caption{Given a desk scene, CoV correctly identifies multiple items on the desk and reasons about their positions, such as the shelf to the right of the monitor and office chairs nearby.}
  \label{fig:pdf_page3}
\end{figure*}

\begin{figure*}[t]
  \centering
  \includegraphics[page=4, width=\linewidth]{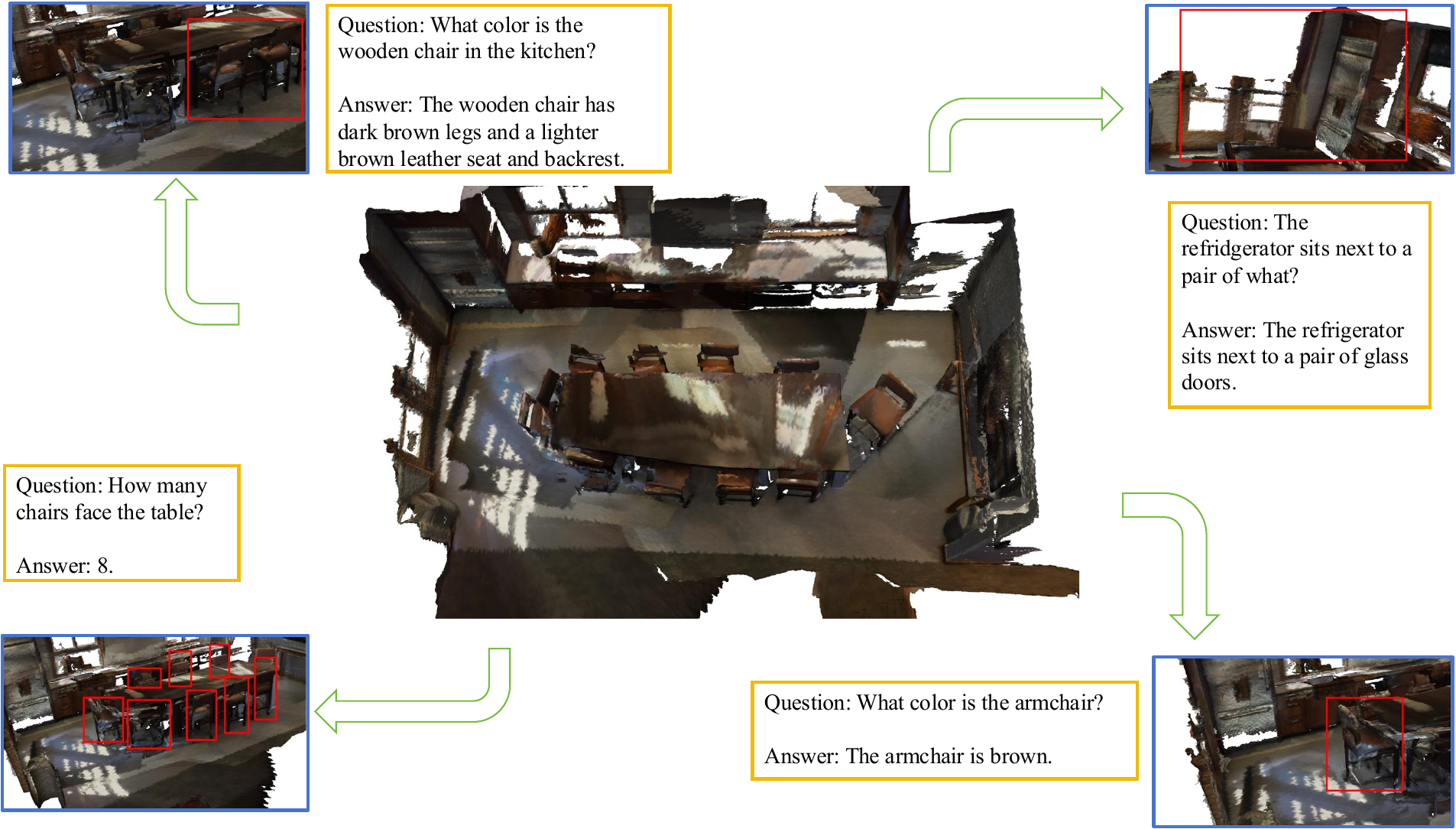}
  \caption{CoV answers layout and appearance questions, including identifying brown-toned wooden chairs, the refrigerator's proximity to glass doors, and counting the number of chairs facing the table.}
  \label{fig:pdf_page4}
\end{figure*}

The qualitative results illustrate CoV’s ability to accurately localize objects, reason about spatial relationships, and align its predictions with the semantics of the question. 
Across diverse indoor scenes—bathroom~\ref{fig:pdf_page1}, classroom~\ref{fig:pdf_page2}, office~\ref{fig:pdf_page3}, and kitchen~\ref{fig:pdf_page4}, CoV consistently identifies relevant details, even when targets are partially occluded or distributed across views. 
These examples demonstrate CoV’s strength in multi-step, viewpoint-aware reasoning.

\clearpage

\end{document}